\documentclass{article}

    \PassOptionsToPackage{numbers}{natbib}

\PassOptionsToPackage{numbers}{natbib}

     \usepackage[final]{neurips_bdl2019}

\usepackage[utf8]{inputenc} 
\usepackage[T1]{fontenc}    
\usepackage{hyperref}       
\usepackage{url}            
\usepackage{booktabs}       
\usepackage{amsfonts}       
\usepackage{nicefrac}       
\usepackage{microtype}      
\usepackage{graphicx}
\usepackage{multirow}
\usepackage{subfigure}
\usepackage{caption}
\usepackage{comment}
\usepackage{amsmath}
\usepackage{appendix}
\usepackage{authblk}

\newcommand{\be}{\begin{equation}}
\newcommand{\ee}{\end{equation}}
\newcommand{\mb}{\mathbf}

\usepackage{array}

\title{Deep Probabilistic Models to Detect Data Poisoning Attacks}

\author[]{Mahesh Subedar, Nilesh Ahuja, Ranganath Krishnan, Ibrahima J. Ndiour, Omesh Tickoo}
\affil[]{Intel Labs }

\newcommand{\PreserveBackslash}[1]{\let\temp=\\#1\let\\=\temp}
\newcolumntype{C}[1]{>{\PreserveBackslash\centering}p{#1}}
\newcolumntype{R}[1]{>{\PreserveBackslash\raggedleft}p{#1}}
\newcolumntype{L}[1]{>{\PreserveBackslash\raggedright}p{#1}}
\newcolumntype{M}[1]{>{\centering\arraybackslash}m{#1}}

\DeclareMathOperator{\E}{\mathbb{E}}
\begin{document}
	\maketitle

	\begin{abstract}
		Data poisoning attacks compromise the integrity of machine-learning models by introducing malicious training samples to influence the results during test time. In this work, we investigate backdoor data poisoning attack on deep neural networks (DNNs) by inserting a backdoor pattern in the training images. The resulting attack will misclassify poisoned test samples while maintaining high accuracies for the clean test-set. We present two approaches for detection of such poisoned samples by quantifying the uncertainty estimates associated with the trained models. In the first approach, we model the outputs of the various layers (deep features) with parametric probability distributions learnt from the clean held-out dataset. At inference, the likelihoods of deep features w.r.t these distributions are calculated to derive uncertainty estimates. In the second approach, we use Bayesian deep neural networks trained with mean-field variational inference to estimate model uncertainty associated with the predictions. The uncertainty estimates from these methods are used to discriminate clean from the poisoned samples. 
	\end{abstract}
	
	\subsection*{1. Introduction}
	Data poisoning attacks are security threats introduced in machine learning models during the training phase.  Deep neural networks (DNNs) require a large amount of training data and compute resources to model complex tasks. In many practical scenarios, it is required to crowd-source the data collection or outsource the model training to external entities.  This poses real security challenges, which can result in compromised machine learning systems. 
	Poisoning attacks have been studied in the context of image classification and computer vision \cite{NIPS2018_7849, NIPS2017_6943,DBLP:journals/corr/abs-1708-06733}, denial of service \cite{Rubinstein:2009:AUD:1644893.1644895}, sentiment analysis \cite{Newell:2014:PIA:2666652.2666661} and malware detection \cite{DBLP:journals/corr/abs-1811-09985, DBLP:journals/corr/abs-1804-07933}. Defensive strategies against data poisoning mostly revolve around anomaly detection (see \cite{DBLP:journals/corr/abs-1802-03041,Carlini:2017:AEE:3128572.3140444} and references therein), though other methods exist \cite{DBLP:journals/corr/abs-1902-06531}. 
	
	In this paper, we consider a type of causative attack reported in \citep{DBLP:journals/corr/abs-1708-06733} where the attacker introduces backdoor pattern in the training dataset (poisoned samples) with the intent to compromise trained model. The resulting model continues to have high accuracy on a held-out clean test-set, and is hence not detectable by such means. This attack scenario is applicable for the trained model obtained from unknown entity or for the model training performed from crowd-sourced data. 
	
	
	\textbf{Contributions:} This paper proposes two promising strategies against data poisoning attacks: a) In the first approach, we model the deep features of a DNN using parametric probability density functions and statistically measure input uncertainty to detect the poisoning attack. b) In the second approach, we use model uncertainty obtained from Bayesian deep neural networks to detect the poisoning attack. Our results show these strategies are effective in detecting data poisoning attacks.

	\subsection*{2. Problem Statement and Approach}
	\label{sec:approach}
	The threat model assumed here is the one presented by \citet{DBLP:journals/corr/abs-1708-06733}. Per this model, the attacker introduces backdoor samples into the training data, $D_{train}$, in such a manner that the accuracy of the resulting trained model measured on a held-out validation set does not reduce relative to that of an honestly trained model. Further, for inputs containing a \emph{backdoor trigger}, the output predictions will be different from those of the honestly trained model. We propose, in this work, two probabilistic methods to derive uncertainty estimates from the compromised deep-network such that lower uncertainty values are obtained for clean samples while higher values are obtained for backdoored or poisoned samples, thereby providing a mechanism to flag such backdoored inputs as anomalous or abnormal during inference. These methods are as follows:

	\paragraph{Probabilistic modeling of deep features (DeepFeatures):} 
	Suppose we have a deep network trained to recognize samples from $N$ classes, $\{C_k\}, k=1, \ldots, N$. Let $f_i(\mb{x})$ denote the output at the $i^{th}$ layer of the network. Our approach consists of fitting class-conditional probability distributions to the features of a DNN, once training is completed. By fitting distributions to the deep features induced by training samples, we are effectively defining a generative model over the deep feature space. At test time, the log-likelihood scores of the features of a test sample are calculated with respect to these distributions. Details of the approach can be found in \cite{ahuja2019probabilistic}. We show that these scores can be used to discriminate clean samples (which should have high likelihood) from poisoned samples (which should have low likelihood). 

	\paragraph{Bayesian Neural Networks (BNN):} 
	We use mean-field variational inference (MFVI) to learn the posterior over weights, and train the model by specifying the prior and approximate posterior as proposed by ~\citet{krishnan2019moped} for efficient and scalable MFVI. We use model uncertainty to distinguish between clean and poison samples. We measure the model uncertainty using Bayesian active learning by disagreement (BALD) \cite{houlsby2011bayesian}, which quantifies mutual information between parameter posterior distribution and predictive distribution.


	\subsection*{3. Experiments and Results}
	\label{sec:Results}
	\textbf{Experimental Setup:} We use MNIST~\cite{lecun1998gradient} and CIFAR10~\cite{krizhevsky2009learning} datasets to study the backdoor data poisoning attack on image classification task. The training set, $D_{train}$, contains all the original clean samples, $D_{train}^{clean}$, along with additional backdoored (BD) training samples, $D_{train}^{BD}$, i.e 
	\begin{equation*}
	    D_{train} = D_{train}^{clean} \cup D_{train}^{BD}
	\end{equation*}We also have a clean held-out validation set, $D_{val}^{clean}$, from which we generate additional backdoored samples, $D_{val}^{BD}$, which will used to measure the effectiveness of the attack and that of our proposed defense. The attack patterns used are the same as those in \citep{DBLP:journals/corr/abs-1708-06733}, in which a four-pixel backdoor pattern is used for MNIST and a 4x4 square pattern is used for CIFAR. For both the datasets, a poisoned sample class label is reassigned to the next class (in a circular count). We study the effect of varying the percentage of poisoned samples in $D_{train}$. 
	We use ResNet-20~\cite{he2016deep} architecture for experiments with CIFAR10 dataset, and a simple convolutional neural network (SCNN) architecture consisting of two convolutional layers followed by two dense layers  for experiments with MNIST dataset.
	
	In the DeepFeatures (DF) approach, DNN models are trained on the $D_{train}$ training set and the parameters of fitted density function are estimated from half of the $D_{val}^{clean}$ held-out set (since we assume no access to the training data). The performance is measured using the remaining half of $D_{val}^{clean}$ (i.e samples not used for fitting the distribution) as well as $D_{val}^{BD}$.

In the Bayesian neural networks (BNN) approach, we  approximate the weight posteriori in variational layers with mean-field Gaussian distribution using Flipout~\cite{wen2018flipout}. The weight prior and approximate posterior for MFVI are specified based on the weights obtained through maximum likelihood estimation from $D_{train}$ using the efficient training approach presented in ~\cite{krishnan2019moped}. During inference phase, predictive distributions are obtained by performing multiple stochastic forward passes over the network while sampling from posterior distribution of the weights (40 Monte Carlo samples in our experiments). We evaluate the model uncertainty with Bayesian active learning by disagreement (BALD) \cite{houlsby2011bayesian} (Equation~\ref{eq:mutual information}) for $D_{val}
^{BD}$ and $D_{train}^{BD}$  test sets.

\textbf{Results:} We demonstrate the effectiveness of poisoning attacks on DNNs and discuss the proposed defense mechanisms against these attacks. The experiments included different percentages of backdoor samples included in the training set. In Table~\ref{tab:Acc}, DNN accuracies are reported for both clean and poisoned datasets. Note that for a poisoned sample, the classification outcome is considered `correct' if it matches the target poisoned label, not the original clean label. Thus, high accuracy on the poisoned dataset indicates that the poisoning attack (with backdoor patterns) has been successful in making the network misclassify the poisoned set while maintaining high accuracy for the clean set.  

For the DeepFeatures (DF) method, if the log-likelihood scores are used to perform classification instead of the softmax scores, the accuracies for the poisoned set are significantly lower than those for the clean set. This indicates that the poisoned samples are not being misclassified as intended by the attacker which suggests that the deep-features themselves are remaining resilient to the poisoning attacks tested here.

\begin{center}
	\begin{table}
		\small
		\renewcommand{\arraystretch}{1.2}
		\begin{tabular}{ C{2cm}|M{1.2cm}|C{1.2cm}| C{9mm} C{9mm} C{9mm} C{9mm} C{9mm} C{9mm} } 
	&		\multicolumn{2}{r}{\textbf{\% of backdoor samples}} &  0 & 10 & 20 & 30 & 40 & 50 \\\hline
\multirow{6}{*}{MNIST} & \multirow{3}{*}{Clean} & DNN & 99.21 & 99.28 & 99.26 & 99.35 & 99.32 & 99.24  \\
&  & DF & - & 99.04 & 98.74 & 98.94 & 99.04 & 98.97 \\ 
&  & BNN & - & 99.56 & 99.64 & 99.56 & 99.51 & 99.5 \\\cline{2-9}
& \multirow{3}{*}{Poisoned} & DNN & - & 98.78 & 98.87 & 99.16 & 99.12 & 99.29 \\
&  & DF & - & 35.54 & 17.55 & 11.87 & 7.32 & 20.54 \\
&  & BNN & - & 99.54 & 99.44 & 99.5 & 99.59 & 99.43 \\\hline
\multirow{6}{*}{CIFAR10} & \multirow{3}{*}{Clean} & DNN & 88.9 & 88.36 & 88.23 & 87.39 & 87.87 & 88.74 \\
&  & DF & - & 88.16 & 88.32 & 88.20 & 88.95 & 88.26 \\
&  & BNN & - & 89.48 & 89.63 & 89.80 & 90.00 & 90.05  \\\cline{2-9}
& \multirow{3}{*}{Poisoned} & DNN & - & 84.08 & 81.95 & 86.4 & 86.94 & 88.29 \\
&  & DF & -  & 69.66  & 64.37  & 75.00  & 75.86  & 80.52 \\
&  & BNN & - & 88.30 & 88.96 & 88.82 & 90.02 & 90.05 \\\hline\hline
\end{tabular}
	\caption{\small Classification accuracies for MNIST and CIFAR10 datasets on Clean (held-out) and Poisoned (backdoor) samples. The backdoor attack is successful in compromising the DNN model by providing similar accuracies for clean and poisoned test samples. DF method is successful in dropping the classification accuracy of the  poisoned data flagging a compromised model. BNN is not effective in modeling this attack as it fits well with training distribution, which includes both clean and poisoned samples. }
	\label{tab:Acc}
\end{table}
\end{center}
\begin{table}
	\small
	\begin{center}
		\renewcommand{\arraystretch}{1.3}
		\begin{tabular}{ C{1.2cm}|C{1cm}| C{1cm}  C{1cm}  C{1cm}  C{1cm}  C{1cm}}
			\multicolumn{2}{c|}{\textbf{\% of backdoor samples}} & 10 & 20 & 30 & 40 & 50 \\\hline
			\multirow{3}{*}{MNIST} & DNN & 83.18 &	81.58 &	68.30 &	54.31 &	53.14 \\
			& DF  & 99.43 &	99.70 &	99.91 &	99.96 &	99.96 \\
			&  BNN & 97.47 & 86.24 & 72.90 &	58.46 &	46.46 \\\hline
			\multirow{3}{*}{CIFAR10}  & DNN & 40.62 & 64.16 & 36.39 & 38.68 & 41.99 \\
			&  DF & 91.87 &	74.58 &	90.94 &	91.81 &	85.17 \\
			&  BNN & 92.24 &	82.03 &	70.95 &	58.89 &	49.22 \\
			\hline
			\hline
		\end{tabular}
		\caption{\small AUPR scores for the DNN, DF and BNN models on MNIST and CIFAR-10 datasets. AUPR scores are calculated using binary classification of clean and poisoned samples. DF and BNN methods show significantly higher scores than the deterministic DNN models.}
		\label{tab:AuPR}
	\end{center}
\end{table}
\vspace{-28pt} 
In the case of Bayesian neural networks (BNN) with MFVI, the accuracies for both the clean and poisoned samples are high indicating the model is not able to differentiate poisoned and the clean datasets. These results are expected since the model is trained on $D_{train}$ set resulting in model learning the distribution of poisoned samples as well. In the case DF method, since the distribution of clean held-out test set are used to fit the distribution, it has an anchor point to differentiate clean from the poisoned samples. In order for BNN to be able to detect the data poisoning attacks, they need to be trained on the held-out clean training set, so that the model can differentiate the distributions of clean and poisoned samples.

In the next set of experiments, we investigate the effectiveness of using the uncertainty methods from DF and BNN methods to distinguish between clean and poisoned samples. The task is set-up as a binary classification task whose performance is evaluated using the precision-recall curve and the area under it (AUPR). In Table~\ref{tab:AuPR}, we present the  AUPR values for both DF and BNN methods. Both DF and BNN methods provide better AUPR values than the deterministic DNN mdoel. BNN performs better at lower percentage of backdoor samples, but with more poisoned samples the BNN model learns the poisoned data distribution resulting in lower AUPR for data poisoning detection.  

\subsection*{4. Discussion}
\label{sec:conc}
The success of the data poisoning attack with simple backdoor patterns (hardly noticeable) show the real threat associated with these attacks on the machine learning models. The two approaches presented here as defense mechanism have the potential to be able to detect the backdoor data poisoning attacks by leveraging uncertainty estimates. Since the uncertainty measures obtained from these two methods are complementary, we will explore ways to combine these methods to model distribution over features and distribution over weights as part of our future work. We presented a potential research thread for these type of poisoning attacks to the research community at large, and we will continue to further investigate these attacks in light of these studies. 

\bibliographystyle{apalike}
\bibliography{BDL_2019}

	\begin{appendices}
	\section{Appendix}
	\label{sec:append}

	\subsection{Uncertainty Quantification}
	
	Given training dataset $D=\{x,y\}$ with  inputs $x = \{x_1, . . . , x_N\}$ and their corresponding outputs $y = \{y_1, . . . , y_N\}$, predictive distribution is obtained through multiple stochastic forward passes through the network while sampling from the posterior of weights $p(w\,|\,D)$ through Monte Carlo estimators.  Equation~\ref{eq:pred_dist} shows the predictive distribution of the output $y^*$ given new input $x^*$: 
    \begin{equation}
    \begin{gathered}
    p(y^*|x^*,D) = \int p(y^*|x^*,w)\,p(w\,|\,D) dw \\
    p(y^*|x^*,D) \approx \frac{1}{T} \sum_{i=1}^{T}p(y^*|x^*,w_i)~,~~~w_i\sim p(w\,|\,D)
    \end{gathered}
    \label{eq:pred_dist}
    \end{equation}
    where, $T$ is number of Monte Carlo samples.
    
    	We estimate the model uncertainty in Bayesian deep neural network using Bayesian active learning by disagreement (BALD) \cite{houlsby2011bayesian}, which quantifies mutual information between parameter posterior distribution and predictive distribution. 
    
	\begin{equation}
	BALD := H(y^*|x^*, D)-\E_{p(w|D)}[H(y^*|x^*, w)]
	\label{eq:mutual information}
	\end{equation}
	where $H(y^*|x^*, D)$ is the predictive entropy as shown in Equation~\ref{eq:pred_entropy}. Predictive entropy captures a combination of input uncertainty and model uncertainty~\cite{gal2016uncertainty}.
	\begin{equation}
	H(y^*|x^*, D):=-\sum_{i=0}^{K-1}p_{i\mu} * log\,p_{i\mu}\\
	\label{eq:pred_entropy}
	\end{equation}
	where $p_{i\mu}$ is predictive mean probability of $i^{th}$ class from $T$ Monte Carlo samples, and $K$ is total number of output classes.
	
	\subsection{Additional Results}
	
	\begin{table}[h]
	\small
	\begin{center}
		\renewcommand{\arraystretch}{1.3}
		\begin{tabular}{ C{1.2cm}|C{1cm}| C{1cm}  C{1cm}  C{1cm}  C{1cm}  C{1cm}}
			\multicolumn{2}{c|}{\textbf{\% of backdoor samples}} & 10 & 20 & 30 & 40 & 50 \\\hline
			\multirow{3}{*}{MNIST} & DNN & 83.60 &	83.31 &	73.41 &	57.65 &	51.66 \\
			& DF  & 99.67 & 99.83 & 99.96 & 99.98 & 99.98 \\
			&  BNN & 83.22 & 69.94 & 61.91 & 55.75 & 51.55 \\\hline
			\multirow{2}{*}{CIFAR10} & DNN & 53.95 &	71.24 &	51.09 &	51.02 &	55.09 \\
			&  DF & 95.75 & 85.09 & 95.57 & 95.78 & 91.96 \\
			&  BNN &  58.64  & 55.23  & 53.03  & 50.34  & 50.13 \\
			\hline
			\hline
		\end{tabular}
		\label{tab:AuROC}
		\caption{\small AUROC scores for the DNN, DF and BNN models on MNIST and CIFAR-10 datasets. AUROC scores are obtained from the binary classification of clean and poisoned datasets.}
	\end{center}
\end{table}
	\end{appendices}
\end{document}